\documentclass[runningheads]{llncs}
\usepackage[T1]{fontenc}
\usepackage{amsmath}
\usepackage{graphicx}
\usepackage{hyperref}       
\usepackage{multirow}
\usepackage{graphicx}
\usepackage{cleveref}
\usepackage{subcaption}
\usepackage{wrapfig}
\usepackage{array}
\usepackage{booktabs}%
\usepackage{afterpage}
\usepackage{comment}
\usepackage{cite}
\usepackage{longtable}
\usepackage{amssymb}
\usepackage{xcolor}
\definecolor{myhighlight}{RGB}{86, 161, 67} 
\begin{document}
\title{Which LIME should I trust? Concepts, Challenges, and Solutions}
%
%
\author{Patrick Knab\inst{1} \and
Sascha Marton\inst{2} \and
Udo Schlegel\inst{3,4} \and
Christian Bartelt\inst{1}}
\authorrunning{P. Knab et al.}
%
\institute{TU Clausthal, Germany \and
University of Mannheim, Germany \and
Ludwig-Maximilians-Universität München, Germany \and
Munich Center for Machine Learning (MCML), Germany}
\maketitle              
\begin{abstract}
As neural networks become dominant in essential systems, Explainable Artificial Intelligence (XAI) plays a crucial role in fostering trust and detecting potential misbehavior of opaque models. 
LIME (Local Interpretable Model-agnostic Explanations) is among the most prominent model-agnostic approaches, generating explanations by approximating the behavior of black-box models around specific instances.
Despite its popularity, LIME faces challenges related to fidelity, stability, and applicability to domain-specific problems. 
Numerous adaptations and enhancements have been proposed to address these issues, but the growing number of developments can be overwhelming, complicating efforts to navigate LIME-related research.
To the best of our knowledge, this is the first survey to comprehensively explore and collect LIME’s foundational concepts and known limitations.
We categorize and compare its various enhancements, offering a structured taxonomy based on intermediate steps and key issues. 
Our analysis provides a holistic overview of advancements in LIME, guiding future research and helping practitioners identify suitable approaches. Additionally, we provide a continuously updated interactive website, \href{https://patrick-knab.github.io/which-lime-to-trust/}{Which LIME Should I Trust?}, offering a concise and accessible overview of the survey.

\keywords{LIME  \and XAI \and Survey.}
\end{abstract}
\section{Introduction}

\textbf{Artificial intelligence (AI)} has transitioned from a futuristic concept to an integral part of our daily lives in the last decade. 
As computing power continues to become more affordable, intelligent models can increasingly be leveraged to enhance process efficiency through automation, elevate user experiences, and enable innovative solutions across diverse sectors such as healthcare, finance, and transportation  \cite{e23010018, Zhou_2021}.
Despite its impressive capabilities, the decision-making processes, primarily driven by neural networks (NNs), remain opaque and challenging to interpret.
This lack of transparency challenges trust, accountability, and ethical considerations in AI applications. 
Therefore, users and stakeholders increasingly demand explanations for AI decisions, particularly in high-stakes areas such as medical diagnoses, loan approvals, and legal judgments. 
These explanations have the goal of helping the user (explainee \cite{MILLER20191}) to build trust in these systems and understand the rationale behind specific outputs \cite{e23010018, BARREDOARRIETA202082}.

\textbf{Explainable AI (XAI)} aims to make AI models more interpretable, addressing the challenges posed by opaque decision-making. 
The concept of interpretability has been widely applied across various domains and to different types of opaque models, such as convolutional neural networks (CNNs) in the imagery domain and transformer-based architectures for textual modalities \cite{9233366, s23020634, molnar2020interpretable, schwalbe2024comprehensive}.
However, because each problem presents unique characteristics, there is a need for diverse explainability techniques \cite{8807299}. 
For instance, one might seek to approximate the behavior of a model \cite{lime}, while in other cases the goal might be to explain the intrinsic properties of the model itself \cite{grad-cam}.
Moreover, developing or extending explanatory approaches to meet new challenges becomes essential as models grow in complexity. 
This could involve incorporating additional data modalities or enhancing the robustness of existing methods to handle more sophisticated architectures or specific data characteristics.

\textbf{LIME} (Local Interpretable Model-agnostic Explanations) \cite{lime} has become one of the most widely adopted techniques in the XAI domain, offering local explanations for complex models by providing insights into how individual decisions are made. 
Despite its popularity, LIME faces several challenges, including instability \cite{glime, baylime, LIME-SUP, Kernel-LIME, alime, GMM-LIME, Segal}, computational inefficiency \cite{Utkin2020SurvLIMEInfAS}, and limitations in the handling of certain types of data \cite{audiolime, dime}. 
In response, numerous studies have proposed enhancements to address these issues \cite{knab2024dseglime, lslime, DBLP:journals/corr/abs-2012-00093, Visani2020OptiLIMEOL, ExpLIMEable, lime_segment}.
Notwithstanding the numerous enhancements, there is a significant need for a thorough and systematic evaluation of LIME and its variants from a research perspective. 
Our new analysis helps guide future developments and expand its application in various domains. 
In addition, practitioners often face challenges in identifying the most suitable LIME methodology or even recognizing the advancements that have been made. 
A comprehensive overview of these techniques supports their ability to explain their models more effectively.

\subsection{Contribution and Organization}
This work presents a comprehensive and focused review of LIME and its diverse adaptations.
Rather than comparing LIME to other interpretability methods such as SHAP \cite{Shap} or Grad-CAM \cite{grad-cam}, we exclusively examine its modifications and extensions to provide a detailed analysis of its evolution, limitations, and research challenges.
By systematically categorizing existing LIME variations, we bridge key gaps in prior research, particularly regarding stability, robustness, and domain-specific adaptations.
Additionally, this survey serves both researchers and practitioners, offering a structured framework to identify suitable LIME techniques based on data modality (e.g., text, images, tabular data) and application domain constraints (e.g., healthcare).
This study also investigates the challenges and limitations of LIME’s application and development. To address these, we perform a detailed literature analysis, introduce a novel categorization framework, and review advancements aimed at improving LIME’s interpretability and efficiency.

The main contributions of this study are as follows:

\begin{itemize}
\item To the best of our knowledge, this paper presents the first comprehensive survey of LIME-related techniques, synthesizing a wide range of modifications and improvements from the literature.
    
\item We introduce a novel taxonomy that categorizes LIME extensions along two key dimensions: (1) the technical modifications within the LIME framework and (2) the specific issues they address. Based on this taxonomy, we systematically analyze the strengths and limitations of existing methods and provide insights into promising directions for future research.
    
\item Beyond categorization, this survey provides a practical resource by mapping research challenges to LIME techniques, helping researchers identify relevant modifications based on specific properties and practitioners select suitable methods for their applications.
\end{itemize}

Furthermore, we host a \href{https://patrick-knab.github.io/which-lime-to-trust/}{webpage} dedicated to this work that will continue to monitor LIME-related techniques, as illustrated in 
\Cref{lime_webpage}.

\subsection{XAI Categorization}
\label{cate}
In the following, we briefly review the categorization of XAI techniques to provide context for situating LIME within the broader landscape of explainability approaches. 
XAI techniques can be broadly categorized into ante-hoc (sometimes also intrinsic) and post-hoc methods \cite{e23010018, BARREDOARRIETA202082, molnar2020interpretable}. 

\textbf{Ante-hoc} methods focus on building inherently interpretable models from the ground up, such as decision trees, linear regression, and rule-based systems. 
These models are designed to be understandable by default but may lack the predictive power of more complex algorithms like neural networks \cite{molnar2020interpretable}.

In contrast, \textbf{post-hoc} methods aim to explain already trained, opaque models without altering their structure. 
These methods are further divided into two types: \textit{model-specific} and \textit{model-agnostic} approaches. 
Model-specific methods generate explanations tailored to particular types of models, such as activation maps in neural networks, or attention in transformers.

\textbf{Model-agnostic techniques}, like LIME \cite{lime} and SHAP (SHapley Additive exPlanations) \cite{Shap}, can be applied to any machine learning model, making them more versatile. 
Both techniques focus on analyzing the relationship between the model’s inputs and outputs rather than examining the internal workings of the model itself. 
Specifically, LIME explains model predictions by approximating the model locally through an interpretable surrogate model built from perturbations of the input features \cite{lime}, while SHAP quantifies the contribution of each feature to a prediction using Shapley values derived from cooperative game theory \cite{Shap}. 
Explanation methods can further be classified into \textit{local} and \textit{global} explanations. 
Local methods, like LIME, explain individual predictions for individual samples. 
In contrast, global methods, such as partial dependence plots (PDP) and feature importance measures, offer insights into the overall behavior of the model in general or across an entire dataset \cite{molnar2020interpretable}.

Recent advancements in XAI have been surveyed extensively, with broad overviews offering foundational insights into general trends, methodologies, and challenges \cite{e23010018, BARREDOARRIETA202082, SAEED2023110273}. 
In parallel, domain-specific reviews have emerged across medical \cite{9233366, s23020634, 10264883}, time-series data \cite{9895252, rojat2021explainableartificialintelligencexai}, IoT \cite{10158334}, tabular data \cite{9551946}, finance \cite{10373833}, manufacturing \cite{10449717}, and human-centered contexts \cite{10316181}, each underscoring the growing need for interpretable, transparent AI solutions in specialized settings.

\section{Methodology} \label{Methodology}
This section outlines the methodology used to search, identify, and analyze research papers on LIME techniques. To ensure transparency in our literature review process, we detail the steps we followed in accordance with established guidelines. Our methodology follows the structured process recommended by Webster et al. \cite{webster}, complemented by the documentation approach outlined by Brocke et al. \cite{brocke}.

\begin{figure}[t]
  \centering
  \begin{subfigure}[b]{0.49\linewidth} 
    \centering
    \includegraphics[width=\linewidth]{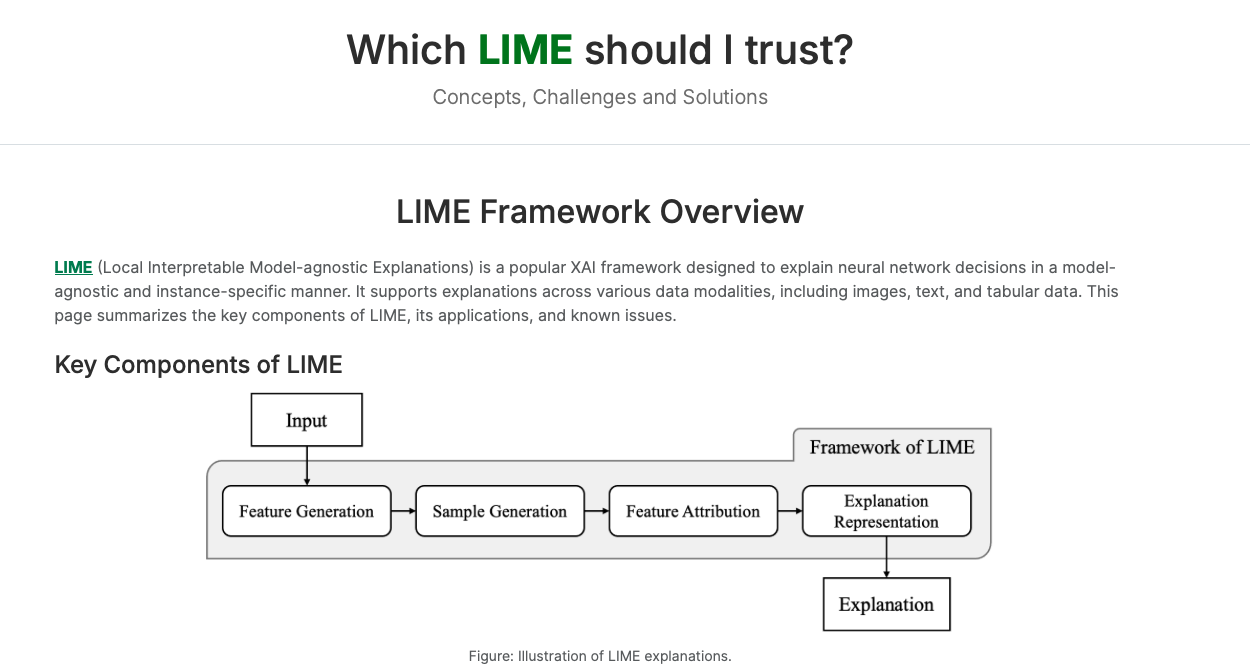}
  \end{subfigure}
  \hfill 
  \begin{subfigure}[b]{0.49\linewidth}
    \centering
    \includegraphics[width=\linewidth]{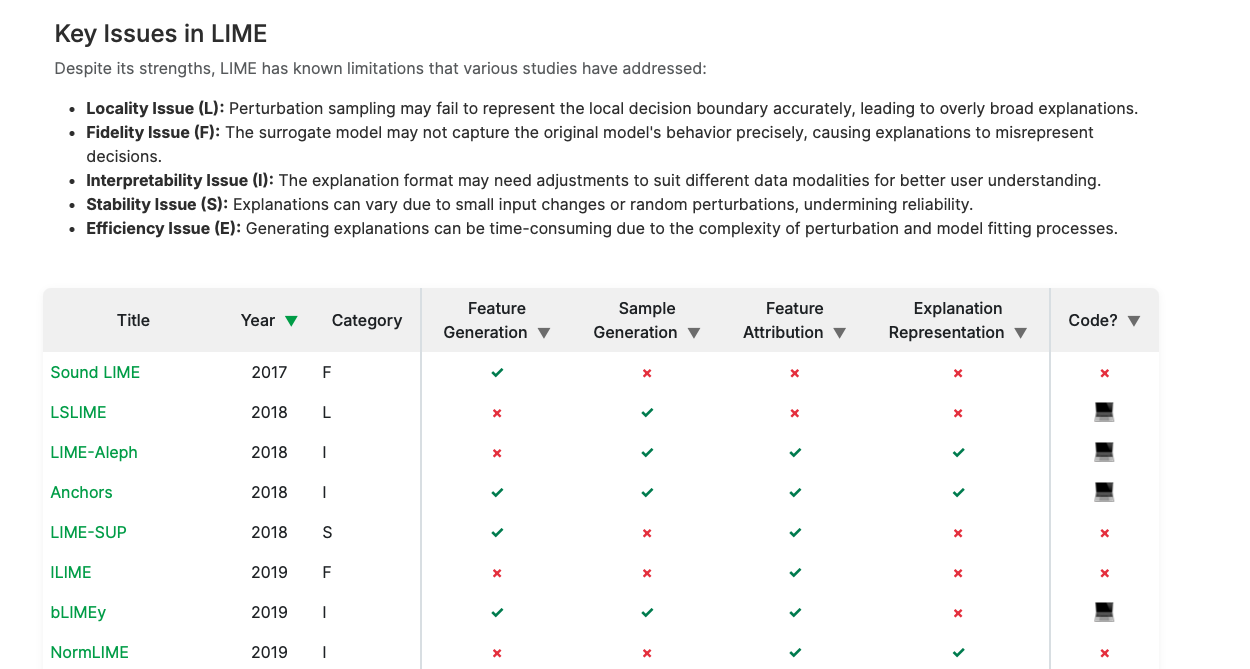}
  \end{subfigure}
  \caption{\textbf{\href{https://patrick-knab.github.io/which-lime-to-trust/}{LIME Webpage}}. This website is designed to monitor and collect new LIME-related techniques continuously. For an ongoing collection of LIME-related methods, please refer to the webpage: \href{https://patrick-knab.github.io/which-lime-to-trust/}{https://patrick-knab.github.io/which-lime-to-trust/}.}
  \label{lime_webpage}
\end{figure}

The literature review began with an extensive exploration of prior research that builds upon and extends the original LIME publication \cite{lime}\footnote{As of January 21st 2025, the publication had been cited in 21,343 papers on ArXiv.}.
Given the vast number of related studies, we refined our selection by incorporating additional keywords such as 'LIME issues', 'LIME improvements', and 'LIME advancements' to systematically narrow the corpus of relevant articles. 
Each selected article was rigorously evaluated based on its quality, applied methodologies, and publication venue.
Furthermore, we included non-peer-reviewed articles from ArXiv, provided they offered novel and pertinent contributions to our research.
Our initial review covered papers published between 2016 and 2025.
To further broaden our search, we examined the references cited in these papers to identify additional relevant works.
Additionally, we focused exclusively on papers that explicitly mention or address LIME limitations and improvements within the LIME framework, recognizing that other works might create model-agnostic explanations outside of this framework \cite{Shap}.
We acknowledge the possibility of overlooking papers that did not align with our selection criteria, given the challenge of reviewing over 20,000 works. 
However, as we maintain a continuous record of LIME advancements, any missed works can be integrated into the ongoing \href{https://patrick-knab.github.io/which-lime-to-trust/}{online overview} (\Cref{lime_webpage}).

To facilitate documentation, we created a concept matrix \cite{webster}. 
It provides comprehensive details on each technique, such as the name, source, the particular problem it addresses, the modality, domain-specific characteristics, a description of any changes made within the LIME framework, the evaluation, and the availability of code for implementation.
This concept matrix built the foundation for the creation of the LIME categorization. 

\section{Fundamentals of LIME} \label{fundamentals}

\noindent
\textbf{Notation.}  
We consider instances from modalities such as time series, images, text, audio, tabular data, or graphs. Let $\mathbf{x} \in \mathcal{X}$ denote an instance, and $\mathbf{y} \in \mathcal{Y}$ its corresponding label.  
For classification, $\mathcal{Y}$ is a set of discrete labels, $\{1, 2, \ldots, C\}$; for regression, $\mathcal{Y} \subset \mathbb{R}$.  
We denote the black-box model as $f: \mathcal{X} \rightarrow \mathcal{Y}$, which outputs a prediction $\hat{y}$ for a given $\mathbf{x}$.

\begin{figure}[t]
  \centering
  \includegraphics[width=\linewidth]{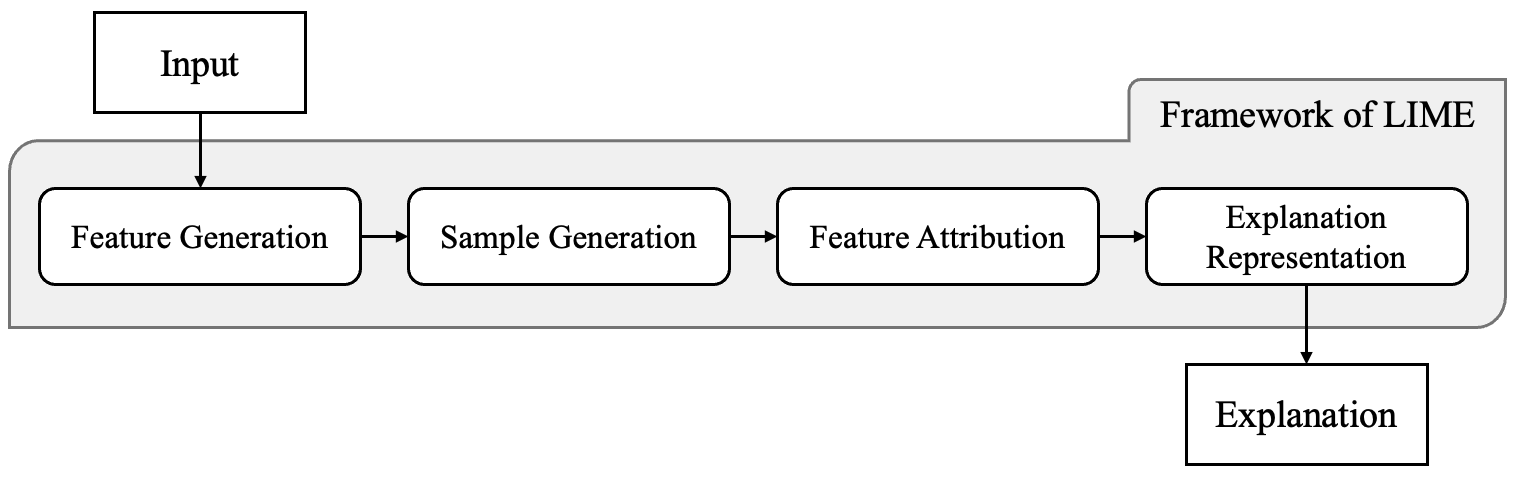}
  \caption{\textbf{Steps of LIME}: The framework operates in four steps: (1) Feature generation: Extract features (e.g., image segmentation). (2) Sample generation: Create perturbed samples around the instance. (3) Feature attribution: Train an interpretable model (e.g., linear) to approximate the complex model locally. (4) Explanation representation: Use the model’s weights to represent feature importance.}
  \label{fig:lime_steps}
\end{figure}

LIME explains the decisions of a neural network $f$ in a \textit{model-agnostic} and \textit{instance-specific} (local) manner, applicable to images, text, and tabular data \cite{lime}. 
Its algorithm follows the structure in \Cref{fig:lime_steps}. 
Through this survey, we will refer to this structure when addressing LIME enhancements and will discuss them in greater detail in \Cref{taxonomy}.

\noindent
\textbf{Feature generation.}
The technique trains a local, interpretable surrogate model $g \in G$ (e.g., linear models or decision trees) to approximate $f$ around an instance $\mathbf{x}$ \cite{lime, LIMEtree}. 
For text or tabular data, minimal preprocessing is applied before computing feature importance scores. 
Preprocessing can include tokenization, lowercasing, and possibly stopword removal for text data. 
Tabular data might involve normalization, handling missing values, or encoding categorical variables. 
A common practice for time series is to generate features based on the time series’ local context, such as rolling window statistics (e.g., moving averages, trends, or differences) that capture the temporal properties in the data. 
In contrast, for imagery data, the $n \times m$ pixels are transformed via segmentation into superpixels \cite{10.5555/3454287.3454867, 6205760, 6976891}, which serve as the features.

\begin{figure}[t]
    \centering
    \begin{minipage}[b]{0.65\textwidth}
        \centering
        \begin{subfigure}[b]{\textwidth}
            \centering
            \includegraphics[width=\textwidth]{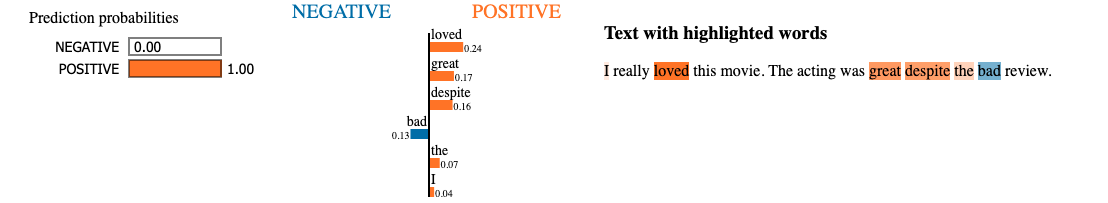}
            \caption{Textual Explanation}
            \label{fig:text}
        \end{subfigure}
        \vspace{0.5cm}
        \begin{subfigure}[b]{\textwidth}
            \centering
            \includegraphics[width=\textwidth]{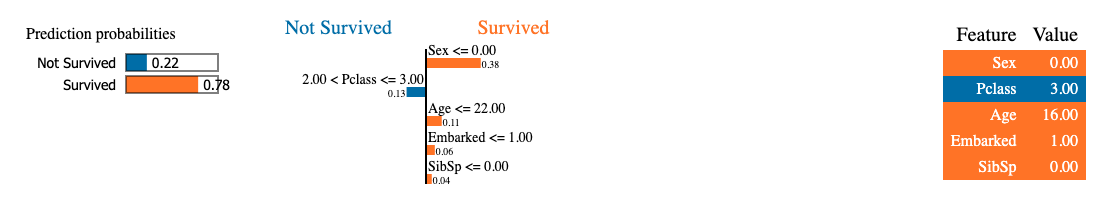}
            \caption{Tabular Explanation}
            \label{fig:tab}
        \end{subfigure}
    \end{minipage}
    \hfill
    \begin{minipage}[b]{0.25\textwidth}
    \centering
        \begin{subfigure}[b]{\textwidth}
            \centering
            \includegraphics[width=0.95\textwidth]{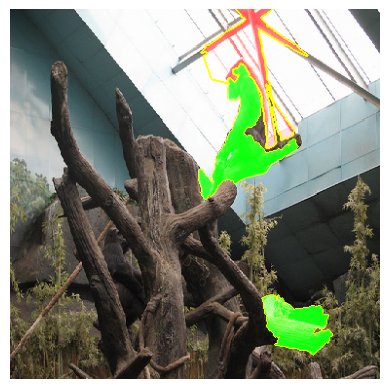}
            \caption{Imagery Explanation}
            \label{fig:image}
        \end{subfigure}
    \end{minipage}
    \caption{\textbf{LIME Exemplary Explanations.} \Cref{fig:text} shows a sentiment classification from a movie review (IMDB dataset \cite{maas-EtAl:2011:ACL-HLT2011}), highlighting words linked to positivity and negativity. \Cref{fig:tab} depicts a young female passenger from the Titanic dataset, with survival probability mainly influenced by her sex and age. \Cref{fig:image} explains an image classified as a gorilla, where green and red superpixels represent positive and negative classification contributions.}
        \label{fig:all_exp}
\end{figure}

\noindent
\textbf{Sample generation.} LIME generates a set of perturbed samples to locally approximate the behavior of a black-box model around a given instance $\mathbf{x} \in \mathcal{X}$. 
Let $\mathbf{z} \in \mathcal{X}$ denote a perturbed instance derived from $\mathbf{x}$ by selectively modifying features while maintaining the underlying structure of the original data. 
The binary vector $\mathbf{z'} \in \{0,1\}^d$ encodes which features are retained ($1$) or replaced ($0$) in the perturbed instance. 
Formally, if the original instance is represented as $\mathbf{x} = [f_1, f_2, \ldots, f_d]$, where $d$ denotes the dimensionality of the feature space (e.g., the number of superpixels in an image, the number of features in tabular data, or the number of tokens in text), then we define $\mathbf{z} = h(\mathbf{x}, \mathbf{z'})$,
where $h(\cdot)$ maps the original instance $\mathbf{x}$ and the binary mask $\mathbf{z'}$ to a perturbed instance $\mathbf{z}$ by retaining or replacing individual features according to:
\[
  z'_i = 
  \begin{cases}
    1 & \text{feature } i \text{ is retained}\\
    0 & \text{feature } i \text{ is replaced}
  \end{cases}.
\]

For \textbf{image data}, the instance $\mathbf{x}$ is segmented into $d$ superpixels $s_1, s_2, \ldots, s_d$ to produce the feature space. 
To generate a perturbed sample $\mathbf{z}$, each superpixel $s_i$ is independently toggled on or off as indicated by $\mathbf{z'}$. If $z'_i = 0$, the pixels in the $i$-th superpixel are replaced by a reference value (e.g., the mean pixel intensity), maintaining the image structure but altering its appearance in selected areas \cite{knab2024dseglime}. 
In the modality type \textbf{tabular data}, each feature $f_i$ is either retained or replaced with a perturbed value sampled from a suitable distribution or set to a reference value \cite{garreau2020looking}. 
This approach preserves the tabular layout but allows variation in feature values. For \textbf{text data}, the instance $\mathbf{x}$ consists of $d$ tokens $w_1, w_2, \ldots, w_d$. 
A perturbed sample $\mathbf{z}$ is generated by retaining or masking/replacing each token based on $\mathbf{z'}$. 
If a token $w_i$ is replaced (i.e., $z'_i=0$), it may be set to a common placeholder such as 'UNK' or removed from the text, thus maintaining syntactic structure while altering semantic content \cite{mardaoui2021analysis}.

By systematically generating perturbed samples $\mathbf{z}$ according to the binary vector $\mathbf{z'}$, LIME explores the local neighborhood of the original instance $\mathbf{x}$, providing the basis for fitting an interpretable model that approximates the model's decision in that locality.

\noindent
\textbf{Feature attribution.}
LIME employs a proximity measure, denoted as $\pi_\mathbf{x}$, to assess the closeness between the predicted outputs $f(\mathbf{z})$ and $f(\mathbf{x})$, which is fundamental in assigning weights to the samples (similarity measurement). 
In the standard implementation of LIME, the kernel $\pi_\mathbf{x}(\mathbf{z})$ is defined as follows:
\[
\pi_\mathbf{x}(\mathbf{z'}) = \exp\left(-\frac{D(\mathbf{x'},\mathbf{z'})^2}{\sigma^2}\right), 
\]
Where $\mathbf{x'}$ is a binary vector, all states are set to 1, representing the original instance $\mathbf{x}$. 
$D$ represents the $L2$ distance (can be substituted with a different distance metric depending on the specific context), given by $D(\mathbf{x'}, \mathbf{z'}) = \sqrt{\sum_{i=1}^{n} (\mathbf{x'}_i - \mathbf{z'}_i)^2}$ and $\sigma$ being the width of the kernel. 
Subsequently, LIME trains a linear model (Surrogate Model), minimizing the loss function $\mathcal{L}$, which is defined as:
\[
\mathcal{L}(f, g, \pi_\mathbf{x}) = \sum_{\mathbf{z}, \mathbf{z'} \in \mathcal{Z}} \pi_\mathbf{x}(\mathbf{z}) \cdot (f(\mathbf{z}) - g(\mathbf{z'}))^2.
\]
In this equation, $\mathbf{z}$, and $\mathbf{z'}$ are sampled instances from the perturbed dataset $\mathcal{Z}$, and $g$ is the interpretable model being learned \cite{lime, glime}. 

\noindent
\textbf{Explanation Representation.}
The model's interpretability comes from the coefficients of $g$, which indicate each feature's influence by their magnitude and sign. 
Explanation communication varies by modality. 
For text, keywords or phrases are emphasized (see \Cref{fig:text}); for tabular data, important features or columns are highlighted (see \Cref{fig:tab}); and for images, the $n$ most influential superpixels (positive or negative) are visually marked (see \Cref{fig:image}).

\newpage
\section{LIME Categorization} \label{taxonomy}
This section constitutes this survey's principal component, wherein we comprehensively elucidate our categorization taxonomy. The first dimension addresses the LIME-specific issues referenced in \cref{categorization}, while \cref{Implementation} delineates the second dimension by succinctly discussing the implementation particulars of the proposed methodologies within the previously established LIME substeps as illustrated in \Cref{fig:lime_steps}. This section wraps up with an overview of LIME techniques with specific modalities and domains for practitioners in \Cref{domains}.

\subsection{Issue Categorization} \label{categorization}
The model-agnostic property allows LIME to handle various machine-learning models effectively. 
Consequently, it has been successfully applied to complex models such NNs, CNNs\cite{diagnostics13111932}, Long-Short-Term-Memory (LSTMs) \cite{foods11142019} networks, and transformer architectures \cite{knab2024dseglime}, as well as decision trees and random forests \cite{DBLP:journals/corr/abs-2012-00093}. Also, its built-in interpretation representation for textual and tabular data makes it easy to use with minimal modifications, as demonstrated in \Cref{fig:all_exp}.
Therefore, the technique has found numerous useful applications across various domains, including healthcare, finance, and manufacturing \cite{knab2024interpretingoutlierstimeseries, https://doi.org/10.1002/ima.23012, diagnostics13111932, 9850272, 9529166, 10440027, Perez-Castanos2024}.

However, the technique is not flawless. Consequently, various issues have emerged over the years that have been identified and tackled by several studies. In the subsequent part, we will group and define these issues into five distinct categories, constituting the first dimension of our terminology. 

\begin{itemize}
    \item \textbf{Locality Issue (L)}: The explanations may not be sufficiently specific to the instance being explained if the perturbed data points used to create the surrogate model do not adequately represent the local decision boundary \cite{Sangroya2020GuidedLIMESS, Bramhall2020QLIMEAQL, audiolime, clime}.

    \item \textbf{Fidelity Issue (F)}: The surrogate model used by LIME may not accurately capture the behavior of the original model, leading to explanations that do not fully reflect the original model's decision-making process \cite{baylime, LIMEtree, Visani2020OptiLIMEOL, ILIME, GMM-LIME, meLIME}. Fidelity and locality are closely linked. The problem of fidelity may arise not only from locality issues, such as when the sampling process fails to generate meaningful perturbed data points but also from other factors, such as an overly simplistic surrogate model or inadequate feature representations.

    \item \textbf{Interpretability Issue (I)}: The explanation representation may not represent the model's decision in a well-interpretable way by users or need adaptions due to varying modalities \cite{ahern2019normlime, ExpLIMEable, ExplainExplore,9811416, blime}. 

    \item \textbf{Stability Issue (S)}: The explanations provided by LIME can vary significantly due to minor changes in the input data, perturbation process, sampling, repeated runs, or the underlying model, resulting in inconsistent and unreliable outcomes. Such behavior undermines confidence in the XAI technique or the model to be explained, as the explainee does not know from which part the instability originates in case of doubt \cite{baylime, BMB-LIME, glime, Segal, LIME-SUP, alime, meLIME}. 

    \item \textbf{Efficiency Issue (E)}: The time required to generate explanations can be significant due to the steps involved in perturbation generation, obtaining model predictions, and fitting the surrogate model \cite{AttentionLIME, Utkin2020SurvLIMEInfAS, KOVALEV2020106164}. 
\end{itemize} 

The issues presented should not be considered independently of each other. 
For example, increasing locality can negatively impact efficiency, but decreasing efficiency can positively affect stability. 
Additionally, many studies discussed here do not focus solely on one issue but address several simultaneously.

\subsection{Implementation Categorization} \label{Implementation}
We already described LIME's functionality in \Cref{fundamentals} that we split into four main parts. 
These substeps contain the essence of the technology (\Cref{fig:lime_steps}), feature (1) and sample generation (2), training an interpretable surrogate model based on the samples (3), and presenting the local explanation with the interpretable model (4), such as its coefficients. 
The LIME-dependent works we cover in this work adapt at one or multiple points of this pipeline; we categorize the techniques with this scheme to get an overview of how different approaches adjust the framework, and we define them more precisely in this subsection. 
\Cref{tab:short_papers} summarizes the techniques, organized by topic, with check marks indicating modifications in the four main components of LIME, which are represented as the table’s columns. This provides researchers with a quick reference to specific LIME variants and their technical properties, making it easier to identify relevant approaches for further development or comparison with similar work.

\afterpage{%
\renewcommand{\arraystretch}{1.0}
\begin{table}[h]
    \centering
    \tiny
    \caption{\textbf{LIME Techniques Categorization.} This table lists all the covered LIME techniques and their associated issues and indicates with a checkmark the point at which the technique has adaptations within the LIME framework.}
    \vspace{5pt}
    \resizebox{0.95\textwidth}{!}{
    \begin{tabular}{>{\raggedright\arraybackslash}p{0.25\linewidth} >
    {\raggedright\arraybackslash}p{0.06\linewidth} >
    {\centering\arraybackslash}p{0.16\linewidth} >
    {\centering\arraybackslash}p{0.16\linewidth} >
    {\centering\arraybackslash}p{0.16\linewidth} >
    {\centering\arraybackslash}p{0.16\linewidth}}
        \toprule
         \textbf{LIME-Technique} & \textbf{Issue} & \textbf{Feature Generation}& \textbf{Sample Generation} & \textbf{Feature Attribution} & \textbf{Expl. Representation}\\
        \midrule

        Kernel-LIME \cite{Kernel-LIME} & L & - & - & \checkmark & -\\
        \midrule
        C-LIME \cite{clime}& L & -& \checkmark& -&-\\
        \midrule
        LSLIME \cite{lslime} & L & - & \checkmark & - & -\\
        \midrule
        QLIME \cite{Bramhall2020QLIMEAQL} & L & - & - & \checkmark & -\\
        \midrule

        ILIME \cite{ILIME}& F & -& -& \checkmark&-\\
        \midrule
        LIMEtree \cite{LIMEtree} & F & - & - & \checkmark & -\\
        \midrule
        Sound LIME \cite{Mishra2017LocalIM} & F & \checkmark & - & - & -\\
        \midrule
        MPS-LIME \cite{mps_lime} & F & - & \checkmark & - & -\\
        \midrule
        US-LIME \cite{US-LIME} & F & - & \checkmark & \checkmark & -\\
        \midrule
        TS-MULE \cite{ts_mule}& F & \checkmark & - & - & -\\
        \midrule
        SS-LIME \cite{SSLIME}& F & - & \checkmark & \checkmark & -\\
        \midrule
        LORE \cite{lore}& F & - & \checkmark & - & \checkmark\\
        \midrule
        LIME-Aleph \cite{lime-aleph, 10.1007/978-3-030-43823-4_16} & I&- & \checkmark& \checkmark & \checkmark\\
        \midrule
        Anchors \cite{anchors} & I & \checkmark & \checkmark & \checkmark & \checkmark\\
        \midrule
        bLIMEy \cite{bLIMEy} & I & \checkmark & \checkmark & \checkmark& -\\
        \midrule
        DIME \cite{dime}&  I & \checkmark& \checkmark& \checkmark& \checkmark\\
        \midrule
        Explain Explore \cite{ExplainExplore}& I & -& -&\checkmark& \checkmark\\
        \midrule
        GraphLIME \cite{9811416} & I & \checkmark& \checkmark& \checkmark & -\\
        \midrule
        NormLIME \cite{ahern2019normlime} & I & - & - & \checkmark & \checkmark\\
        \midrule

        $\mathcal{G}$-LIME \cite{LI2023103823} & S & \checkmark & \checkmark& \checkmark& -\\
        \midrule
         DLIME \cite{dlime} & S &\checkmark & \checkmark& \checkmark &-\\
        \midrule
        LIME-SUP \cite{LIME-SUP} & S  & \checkmark & - & \checkmark & -\\
        \midrule
        K-LIME \cite{hall2017machine} & S & \checkmark & - & \checkmark & \checkmark\\
        \midrule
        S-LIME \cite{Zhou_2021} & S & - & \checkmark & - & -\\
        \midrule
        Attention-LIME \cite{AttentionLIME}& E & - & \checkmark & - & -\\
        \midrule
        survLIME \cite{KOVALEV2020106164} & E & - & - & \checkmark & -\\
        \midrule
        survLIME-inf \cite{Utkin2020SurvLIMEInfAS} & E & - & - & \checkmark & -\\
        \midrule

        Specific-Input LIME \cite{specific-input_LIME}& L, F & \checkmark & - & - & -\\
        \midrule
        s-LIME \cite{10.1007/978-3-031-01333-1_9} & L, F & - & \checkmark & - & -\\
        \midrule
        LIMESegment \cite{lime_segment} & L, F & \checkmark & \checkmark & - & -\\
        \midrule
        CBR-LIME \cite{cbr_lime} & L, F & - & \checkmark & \checkmark & -\\
        \midrule
        GuidedLIME \cite{Sangroya2020GuidedLIMESS} & L, I& -& \checkmark& -&-\\
        \midrule
        audioLIME \cite{audiolime} & L, I& \checkmark & - & - & \checkmark\\
        \midrule

        ExpLIMEable \cite{ExpLIMEable}& S, I & \checkmark& \checkmark& -&\checkmark\\
        \midrule
        Sig-LIME \cite{10488390} & S, I & \checkmark & \checkmark & \checkmark & -\\
        \midrule
        gLIME \cite{dikopoulou2021glime} & S, I & - & - & \checkmark & \checkmark\\
        \midrule
    
        ALIME \cite{alime} &S, F& - &\checkmark& \checkmark&-\\
        \midrule
        BayLIME \cite{baylime} & S, F &-&- &\checkmark & -\\
        \midrule
        BMB-LIME \cite{BMB-LIME}& S, F& -& \checkmark& \checkmark&-\\
        \midrule
        SLICE \cite{Bora_2024_CVPR} & S, F&\checkmark & \checkmark& - & -\\
        \midrule
        OptiLIME \cite{Visani2020OptiLIMEOL} & S, F & - & - & \checkmark & -\\
        \midrule
        DSEG-LIME \cite{knab2024dseglime} & S, F & \checkmark & \checkmark & \checkmark & \checkmark \\
        \midrule
        UnRAvEL-LIME \cite{unravel} & S, F & - & \checkmark & - & -\\
        \midrule
        GLIME \cite{glime} & S, F  &-& \checkmark&-& -\\
        \midrule
        GMM-LIME \cite{GMM-LIME}& S, F & -& \checkmark& -&-\\
        \midrule
        MeLIME \cite{meLIME} & S, F & - & \checkmark & \checkmark & -\\
        \midrule
        SEGAL \cite{Segal} & S, L & - & \checkmark & \checkmark & -\\
        \midrule
        B-LIME \cite{blime} & S, F, I &\checkmark & \checkmark & \checkmark & \checkmark\\
        
        \bottomrule
    \end{tabular}
    }
    \label{tab:short_papers}
\end{table}
\clearpage
}
\noindent
\textbf{Feature Generation (1)}: 
This step encompasses all processes that modify the input instance to generate new or transformed features, enhancing their suitability for the explanation process.

\textit{Segmentation-based}:
Given the variety of segmentation techniques, the authors of \cite{bLIMEy} enable individual selection of feature generation methods in their pipeline, similar to the approach in \cite{ExpLIMEable}. With the advancement of foundation models like Segment Anything (SAM) \cite{sam}, Knab et al. integrate these models into their framework to automatically segment images into more interpretable segments for humans \cite{knab2024dseglime}. 
Segmentation variations are not confined to images but extend to preprocessing time series. Several papers investigate segmentation approaches tailored explicitly to time series data \cite{blime, ts_mule, 10488390}. With more meaningful segments, for example, through semantic correlations \cite{lime_segment}, the super features can be covered in some coherent regions, allowing the model's decision to be better approximated. For audio input, the authors of \cite{audiolime} employ source separation to decompose the signal into its sources, such as piano, vocals, or drums, and then segment these audio signals. Sound LIME \cite{Mishra2017LocalIM} further segments the audio signal into temporal, frequency, and time-frequency parts. When dealing with graph data, GraphLIME \cite{9811416} uses the nodes within the graph as features. 

\textit{Clustering-based}:
Other methods apply clustering to the training set to incorporate this knowledge into the process. For instance, \cite{dlime} uses hierarchical clustering to partition the training set into specific clusters, akin to the approach in \cite{hall2017machine}. The authors of \cite{LIME-SUP} propose a similar method but with a supervised partitioning tree for clustering.

\textit{Feature importance-based}:
An et al. \cite{specific-input_LIME} calculate feature importance and generate a partial dependence plot (PDP) in advance, feeding the feature importance scores, split feature importance, and the binary PDP plot into LIME. SLICE \cite{Bora_2024_CVPR} calculates feature importance for feature selection with the estimation of sign entropy. Conversely, the authors of \cite{LI2023103823} first run standard LIME on every sample of the training set, aggregating these explanations to create a global explanation using NormLIME \cite{ahern2019normlime} and Averaged-Importance \cite{vanderlinden2019global} as priors in a Bayesian framework.

\textit{Arbitrary-based}: The authors of the initial LIME pipeline also propose Anchors \cite{anchors}, which create 'anchor' rules by identifying key features and using these rules as explanations; these anchors are linked to features, such as superpixels in images or specific features in tabular data. For multimodal tasks, e.g.,  Visual Question Answering (VQA), DIME \cite{dime} applies the feature generation processes for both modalities (image and text) in parallel.

\noindent
\textbf{Sample Generation (2)}: Generating samples for explanation purposes can be achieved through various methods. Techniques that deviate from the standard LIME process, which typically involves random sampling around an instance, adopt more refined approaches. Many of these methods focus on selective, approximation-based, neighborhood-based, or distribution-based sampling strategies to improve the interpretability and reliability of the explanations.

\textit{Selective-based}: Sangroya et al. \cite{Sangroya2020GuidedLIMESS} proposes a method that selects samples, maximizing coverage criteria while minimizing redundancy. BMB-LIME \cite{BMB-LIME} incorporates uncertainty estimation from BayLIME \cite{baylime} to enrich sample diversity, whereas \cite{clime} excludes anomalous samples that deviate significantly from the local neighborhood. LIME-Aleph \cite{lime-aleph} uniformly selects $n$ instances from a pool of logical representations, while US-LIME \cite{US-LIME} pre-generates samples and then focuses on those near the decision boundary. UnRAvEL-LIME \cite{unravel} employs an uncertainty-driven acquisition function to guide sample selection.

\textit{Approximation-based}: In the work of Li et al. 
\cite{LI2023103823}, they use a modified version of ElasticNet \cite{10.1111/j.1467-9868.2005.00503.x} estimator with \(\ell_1\) and \(\ell_2\) regularization to generate sparse, informative interpretations in Bayesian linear regression. AttentionLIME \cite{AttentionLIME} narrows the search space by leveraging attention weights from the target label, improving sampling efficiency for text data. S-LIME \cite{Zhou_2021} uses Lasso, a modification of Least Angle Regression (LARs) \cite{10.1214/009053604000000067}, to estimate the number of samples required for more stable explanations.

\textit{Neighborhood-based}: DLIME \cite{dlime} utilizes the partitions from the feature generation process and applies KNN to find the closest neighbors and samples from those. MeLIME \cite{meLIME} samples from neighbors with different generators (kernel density, principal component analysis, variational autoencoder, and word2vec). In a graph-fashion style, MPS-LIME \cite{mps_lime} structures the superpixels in an undirected graph; connected nodes are neighbors and search for cliques for sampling. Similar to GraphLIME \cite{9811416}, it also acts in a graph-like structure but applies neighborhood sampling using the proposed N-hop network. LSLIME \cite{lslime} generates instances with GrowingSpheres to find close decision boundaries. s-LIME \cite{10.1007/978-3-031-01333-1_9} generates $n$ equally-weighted instances independently drawn from the underlying distribution. DSEG-LIME \cite{knab2024dseglime} samples from a hierarchical tree, where parent and child segments form the nodes within the tree. Anchors \cite{anchors} generates samples around the test instance to validate the generated anchors and LORE \cite{lore} uses a genetic algorithm. The authors of \cite{cbr_lime} propose an approach by selecting similar instances with a predetermined set of LIME settings for the instance to be explained to achieve better results.

\textit{Distribution-based}: ALIME \cite{alime} samples from a Gaussian distribution with a trained autoencoder. \cite{glime} employs a local and unbiased sampling distribution to ensure the locality property. SEGAL \cite{Segal} has a generative model that samples from the underlying data distribution. GMM-LIME \cite{GMM-LIME} utilizes a Gaussian Mixture Model (GMM) to capture the data's underlying distribution and sample from it. 

\textit{Arbitrary-based}: The authors of \cite{blime} use bootstrapping to sample from time series segments. Sig-LIME \cite{10488390} introduces random noise while sampling time series segments. SLICE \cite{Bora_2024_CVPR} employs adaptive blur in its superpixel replacement strategy. In \cite{dime}, the sample generation process is expanded to incorporate two modalities. The works of \cite{bLIMEy, ExpLIMEable} propose a modular approach, allowing users to switch between various sampling techniques. Sivill and Flach \cite{lime_segment} emphasize the significant impact of perturbed backgrounds on LIME's performance and propose an approach to generate realistic perturbations for time series data. SS-LIME \cite{SSLIME} generates perturbations by substituting words or tokens with semantically similar alternatives, ensuring that the sampled instances preserve contextual coherence.

\noindent
\textbf{Feature Attribution (3)}: This step includes all interpretable surrogate model training modifications by replacing the interpretable model, altering the proximity measurement or kernel, or other alterations. 

\textit{Linear regression modifications}:
\cite{baylime} uses a Bayesian linear regression approach with the optional integration of priors to increase stability.
$\mathcal{G}$-LIME \cite{LI2023103823} utilizes Least Angle Regression (LARs) \cite{10.1214/009053604000000067} to rank the importance of every feature in the explanation over the path of \(\ell_1\)-regularization.
K-LIME \cite{hall2017machine} builds a generalized linear model 

\textit{Replacement of surrogate model}:
LIME-SUP \cite{LIME-SUP} and LORE \cite{lore} replace the model with a tree-based technique. Similarly, \cite{blime} trains a decision forest as its surrogate model. Another tree-based technique is Sig-LIME \cite{10488390}, which uses random forests to uphold non-linear interactions and use them for the explanation. LIMEtree \cite{LIMEtree}, in contrast, uses multi-output regression trees to improve fidelity.
gLIME \cite{glime} implements a graphical most minor absolute shrinkage and selection operator (GLASSO) \cite{epskamp2018tutorial} that produces an undirected Gaussian graph for explanation.
\cite{KOVALEV2020106164} utilizes the Cox proportional hazards model \cite{lin1989robust} to approximate a survival model as the surrogate model. \cite{Utkin2020SurvLIMEInfAS} improves this approach \cite{KOVALEV2020106164} by using the $L_{\infty}$-norm for computing the distance between cumulative hazard functions.  
LIME-Alpeh \cite{lime-aleph} replaces the model for interpretation with an inductive logic programming system that enables the detection of feature relations.
Since decision boundaries are limited by linear regression, BMB-LIME \cite{BMB-LIME} approximates nonlinear boundaries using multivariate adaptive regression splines (MARS) and employs bootstrap aggregation to stabilize the explanation.
In contrast, \cite{Bramhall2020QLIMEAQL} learns non-linear decision boundaries by fitting a quadratic model.

\textit{Weighting modification}:
\cite{dlime} calculates the pairwise distance of the instances within the built clusters and uses them as the weights for the training of the surrogate model.
\cite{alime} uses the embedding of the trained autoencoder for the weighting of the samples used for surrogate model training.
ILIME \cite{ILIME} includes the proximity of the generated samples to the original instance as an influencing function.
OptiLIME \cite{Visani2020OptiLIMEOL} automatically determines the optimal kernel width to balance the explanation stability and fidelity trade-off. Also, \cite{Kernel-LIME} improves the locality of the explanation by an automatic kernel adaption during the surrogate model training.
US-LIME \cite{US-LIME} has a weighting kernel approach that is based on a Gaussian kernel.
Other hyperparameters that can be optimized are treated by \cite{Segal}, which utilizes an adaptive weighting method for additional hyperparameter tuning. ExplainExplore \cite{ExplainExplore} leaves this parameter adjustment dynamically adaptive by the explainee during the explanation process.

\textit{Training of surrogate model}:
\cite{anchors} evaluates and assigns importance to its anchors by measuring their coverage and precision in the predictions. 
\cite{knab2024dseglime} iteratively goes through the feature importance calculation in the built tree structure to construct fine or coarse explanations.
MeLIME \cite{meLIME} uses a local minibatch strategy, deduced by the neighboring approach, to improve the robustness of the training of the surrogate model.  
NormLIME \cite{ahern2019normlime} shifts from local to global explanations by running multiple LIME instances and aggregating them to estimate the global relative importance of the model’s features.
\cite{9811416} trains the model using the Hilbert-Schmidt
Independence Criterion Lasso (HSIC Lasso).
bLIMEy \cite{bLIMEy} provides a modular architecture enabling users to select between different feature attribution techniques as in the previous steps. 
DIME \cite{dime} has separate feature attribution methods to calculate the unimodal importance contributions for each modality by disentanglement and the multimodal contribution.

\noindent
\textbf{Explanation Representation (4)}: The explanation derived from the surrogate model is typically static, visualizing the most significant coefficients as a saliency map for images or highlighting the most critical features within text or tabular data. However, this representation might overlook essential properties needed for better interpretation by the explainee. 

\textit{Expanded explanation representation: }
Anchors \cite{anchors} generate a set of predicates that serve as explanations for the explainee. For the sentiment analysis, the explanation could include rules like 'not bad' indicating positive sentiment, and 'not good' indicating negative sentiment. Similarly, LORE \cite{lore} translates the rule from the decision tree into a logical explanation.
In \cite{knab2024dseglime}, the granularity of the explanation can be controlled by the explainee, allowing visualization of both broad and coarse explanations derived from segmentation hierarchies, shown as feature importance maps for image classification. \cite{glime} produces a graph-based explanation, highlighting feature relationships of nodes and information flows between them. Similarly, \cite{lime-aleph} considers combinations of features and their relationships within an object for explanation. B-LIME \cite{blime} extends heatmap visualization to fit time-series data, indicating which points or sequences in the time series are most important based on the segmented data.
For sound modalities, AudioLIME \cite{audiolime} provides listenable explanations, improving interpretability and broadening the range of explanation representations. NormLIME \cite{ahern2019normlime} transitions the type of explanation from local to global, allowing users to make global assumptions about the behavior of the model. For multimodal expansion, DIME \cite{dime} generates unimodal contribution explanations for text and images, along with a multimodal interaction explanation to explore how the different modalities interact.

\textit{Interactive explanation: }
ExpLIMEable \cite{ExpLIMEable} offers an interactive dashboard to the explainee, allowing the modification of the explanation parameters during the explanation phase to improve the output. Similarly, ExplainExplore \cite{ExplainExplore} provides a dashboard with multiple visual explanations, offering different perspectives to aid interpretability.

\subsection{Domain Categorization} \label{domains}

LIME's fundamental implementation has been utilized for text, tables, and images. Our review of the literature uncovered various specific modes, adaptations for particular domains, and multimodal strategies. \Cref{tab:domains} presents an overview of these approaches, detailing their associated modalities and domains, providing practitioners with a convenient reference for identifying suitable techniques.

\begin{table}[t]
    \centering
    \caption{\textbf{Modalities and Domains of LIME Adaptions.} This table clusters the adaptations covered in this survey into the covered modalities and the additional domain-specific adaptations.}
    \label{tab:domains}
    \vspace{5pt}
    \tiny
    \resizebox{\textwidth}{!}{%
    \begin{tabular}{>{\raggedright\arraybackslash}p{0.1\linewidth} >
    {\raggedright\arraybackslash}p{0.12\linewidth} >
    {\raggedright\arraybackslash}p{0.75\linewidth}}
        \toprule
         \textbf{Modality} &\textbf{Domain}& \textbf{LIME-Techniques}\\
        \midrule
        \multirow{2}{*}{Universal} & Text, Tabular \& Image & LIME \cite{lime}, BayLIME \cite{baylime}, GLIME \cite{glime}, MeLIME \cite{meLIME}, s-LIME \cite{10.1007/978-3-031-01333-1_9}, Anchors \cite{anchors}, $\mathcal{G}$-LIME \cite{LI2023103823}, NormLIME \cite{ahern2019normlime}, OptiLIME \cite{Visani2020OptiLIMEOL}, S-LIME \cite{Zhou_2021}  \\
        \midrule
        \addlinespace
        \multirow{4}{*}{Image} & Universal & DSEG-LIME \cite{knab2024dseglime}, SLICE \cite{Bora_2024_CVPR}, MPS-LIME \cite{mps_lime}, CBR-LIME \cite{cbr_lime}\\ 
                    \addlinespace
                  & Archaeology & LIME-Aleph \cite{lime-aleph, 10.1007/978-3-030-43823-4_16}\\
                  \addlinespace
                  & Healthcare & ExpLIMEable \cite{ExpLIMEable} \\
        \midrule
        \addlinespace
        \multirow{3}{*}{Time Series} & Universal & SEGAL \cite{Segal}, TS-MULE \cite{ts_mule}, LIMESegment \cite{lime_segment}\\
                    \addlinespace
                  & Healthcare & B-LIME \cite{blime}, Sig-LIME \cite{10488390}, C-LIME \cite{clime}\\
        \midrule
        \addlinespace
        
        \multirow{5}{*}{Tabular} & Universal & ILIME \cite{ILIME}, Kernel-LIME \cite{Kernel-LIME}, K-LIME \cite{hall2017machine}, LIME-SUP \cite{LIME-SUP}, survLIME \cite{KOVALEV2020106164}, survLIME-inf \cite{Utkin2020SurvLIMEInfAS}, bLIMEy \cite{bLIMEy}, GMM-LIME \cite{GMM-LIME}, BMB-LIME \cite{BMB-LIME}, LORE \cite{lore}, ExplainExplore \cite{ExplainExplore}, GuidedLIME \cite{Sangroya2020GuidedLIMESS}, LSLIME \cite{lslime}, QLIME \cite{Bramhall2020QLIMEAQL}, Specific-Input LIME \cite{specific-input_LIME}, US-LIME \cite{US-LIME}, LIMEtree \cite{LIMEtree}, ExplainExplore \cite{ExplainExplore}, UnRAvEL-LIME \cite{unravel}
        \\ \addlinespace
                  & Healthcare & DLIME \cite{dlime}, gLIME \cite{dikopoulou2021glime}, ALIME \cite{alime}\\
        \midrule
        \addlinespace
        \multirow{1}{*}{Audio} & Universal & audioLIME \cite{audiolime}, Sound LIME \cite{Mishra2017LocalIM}\\
        \midrule
        \addlinespace
        \multirow{2}{*}{Text} & Universal & AttentionLIME \cite{AttentionLIME} \\
            \addlinespace
                  & Semantic Analysis & SS-LIME \cite{SSLIME}\\
        \midrule
        \addlinespace
        \multirow{1}{*}{Image-Text} & VQA & DIME \cite{dime}\\
        \midrule
         \addlinespace
        \multirow{1}{*}{Graph} & Universal & GraphLIME \cite{9811416}\\
        \bottomrule
    \end{tabular}
    }
\end{table}

As the table illustrates, some implementations are categorized as universal, capable of handling text, tabular, and image data without requiring modifications, and independent of any specific domain. Other approaches are categorized by modality, with some exhibiting domain-specific characteristics, such as those designed for archaeology or healthcare, if the method has been exclusively tested within that domain. Approaches within a particular modality are also considered universal if the authors do not mention any limitations in their work. However, we cannot definitively assess the validity of the claimed universality for each approach, as not all methods have an available code.

\section{Discussion, Opportunities \& Conclusion} \label{i_o}
We conclude this paper with a section addressing key research challenges specific to LIME and outlining potential solutions. This is followed by future research directions related to LIME and finalized with the conclusion.

\subsection{Discussion} \label{lime_problems}

Despite LIME's popularity, a best-practice standard for research and evaluation has not been established. 
This lack of standardized procedures hinders the scientific rigor and broader application of LIME-based techniques. 
In this section, we address the most significant issues that have arisen in LIME research and application without explicitly naming specific papers that have not met these standards. 
By highlighting these problems, we aim to encourage the development of more robust and reproducible LIME-related methodologies, ultimately enhancing the trust and interpretability of model predictions.

\begin{table}[tb]
    \centering
    \tiny
    \caption{\textbf{LIME XAI Evaluation Metrics.} This table categorizes key properties across different areas, such as content, presentation, and user-centered aspects, to guide researchers and practitioners in selecting suitable metrics for evaluating LIME explanations. We adopt the structure of \cite{Nauta_2023}.}
    \vspace{5pt}
    \label{Tabl:eval}
    \resizebox{\textwidth}{!}{%
    \begin{tabular}{>{\raggedright\arraybackslash}p{0.07\linewidth} >
    {\raggedright\arraybackslash}p{0.15\linewidth} >
    {\raggedright\arraybackslash}p{0.9\linewidth}}
        \toprule
         \textbf{Area} &\textbf{Property}& \textbf{Description}\\
        \midrule
        \multirow{25}{*}{\rotatebox{90}{Content}} & Correctness & Reflects how accurately the explanation represents the underlying model’s decision-making process.\\
            \addlinespace
            & Completeness & Measures the extent to which the explanation captures the full behavior of the model, including all relevant aspects that influence its decisions.\\
            \addlinespace
            & Consistency & Evaluates how stable the explanation is across multiple similar instances, ensuring the same explanation is provided when the inputs are only slightly varied or the same.\\
            \addlinespace
            & Continuity & Assesses the degree of similarity between explanations for similar instances, ensuring that small changes in input lead to minor adjustments in the explanation.\\
            \addlinespace
            & Contrastivity & Gauges how well the explanation differentiates the explained instance from others, clarifying why this instance was treated differently than similar cases.\\
            \addlinespace
            & Covariate complexity & Refers to the complexity of the features and their interactions used in the explanation, with simpler explanations typically being more interpretable.\\
            \addlinespace
            & Efficiency & Evaluates the computational resources and time required to generate explanations, which are especially important for real-time applications.\\
            \addlinespace
            & Scalability & Describes how well the explanation method performs as the model or dataset size increases.\\
            \addlinespace
        \midrule
        \addlinespace
        \multirow{15}{*}{\rotatebox{90}{Presentation}} & Compactness & Describes the conciseness of the explanation, focusing on its size and the number of elements included, with smaller explanations generally being more user-friendly.\\
            \addlinespace
            & Composition & Refers to the structure and organization of the explanation, including how the information is presented and whether it follows a logical format.\\
            \addlinespace
            & Confidence & Indicates the inclusion and accuracy of probability information within the explanation, helping users understand the certainty associated with the model’s predictions.\\
            \addlinespace
            & Applicability & Evaluates whether the explanation method can be used across various models or if it's tied to specific models.\\
            \addlinespace
            & Modality Flexibility & Describes whether the explanation method can adapt to different data modalities, such as text, image, or tabular data.\\
            \addlinespace
        \midrule
        \addlinespace
        \multirow{12}{*}{\rotatebox{90}{User}} & Context & Describes how well the explanation aligns with the user’s specific needs, goals, and the context in which it is applied.\\
            \addlinespace
            & Coherence & Refers to how well the explanation fits with the user’s prior knowledge, beliefs, and expectations, ensuring that it makes logical sense from their perspective.\\
            \addlinespace
            & Controllability & Describes the level of interactivity or customization the user has over the explanation, allowing them to explore or adjust different aspects of it.\\
            \addlinespace
            & Complexity & Refers to how easy or complex it is for a user to generate an explanation, including the required computational resources, technical knowledge, and effort involved. \\
        \bottomrule
    \end{tabular}
    }
\end{table}

\textbf{Reproduction Issues.}
We observed a great lack of code availability (50\%, see tool in \Cref{lime_webpage}) that creates significant reproducibility issues, making verifying the authors' contributions difficult. 
This limitation hinders the XAI community's ability to adapt to more advanced LIME frameworks, address known issues, and improve the trust and interpretability of the model's predictions that the explainee seeks to achieve. 

\textbf{Evaluation Practices.}
A direct consequence of the code availability limitation becomes evident when examining the evaluation sections of the covered articles. This lack prevents authors from comparing their LIME-related approaches to previously published techniques for benchmarking. 
As a result, many papers begin their motivation by highlighting a known issue of LIME (see \Cref{categorization}) and often compare their adapted version solely against the standard LIME version. 
However, this approach disregards other related works, a practice that undermines the quality of the research. In contrast to the works that include a comparison with previous LIME-related techniques, these often compare against S-LIME \cite{Zhou_2021} or BayLIME \cite{baylime}, as these provide code for reproducibility and thus enable comparisons.

Another prevalent problem is the selection of evaluation metrics that confirm the stated contributions. This issue is universal within the XAI community and needs to be addressed. The evaluation of the explanations should be divided into qualitative and quantitative evaluations. Many works integrate user surveys for qualitative evaluation, where participants rate XAI-generated explanations. However, there is no standard procedure, with varying sample sizes and differing tasks—some involve interacting with explanations, while others require selecting the best one. We encourage authors to conduct user studies consistently and scientifically, referring to \cite{chromik2020taxonomy}.

Concerning quantitative evaluation, numerous metrics have been proposed to assess various aspects of what makes a good explanation. \cite{10.1145/3583558} provide a comprehensive overview of these techniques aimed explicitly at explainable AI. However, despite the availability of these metrics, their application remains inconsistent. 
Many papers address only a subset of evaluation criteria, creating ambiguity about whether the proposed approaches are truly superior or where they may have limitations, which are not always clearly communicated. 
Regarding this issue, we identified the work of Klein et al. 
\cite{klein2024navigatingmazeexplainableai} highlighting a similar concern, noting that practitioners often struggle to choose the proper method for their specific problem. 
Their work demonstrates how evaluation metrics, when paired with different model architectures, can guide the search for appropriate XAI techniques, aligning with the challenges we address in this paper. 
In \Cref{future_work}, we explore how LIME evaluation can be standardized to address these issues.
In \Cref{Tabl:eval}, we provide an overview of XAI metrics applicable to LIME, which can guide researchers in selecting proper evaluation metrics for their use cases. Here, we adapt the structure proposed by Nauta et al. \cite{Nauta_2023} and add additional metrics found during the literature review.

\subsection{Research Opportunities}  \label{future_work}


\textbf{Automatic LIME Selection.}
The challenge of selecting an appropriate XAI technique has been addressed in the literature \cite{wenzhuo2022-omnixai, DBLP:journals/corr/abs-2009-07896, JMLR:v22:21-0017, 8807299}, as the growing number of methods can make it difficult for users to choose the most suitable one. These studies guide users in applying different techniques based on specific requirements. An advancement in this area is presented in \cite{10.1145/3511808.3557247}, which proposes an automated approach for selecting XAI techniques based on factors such as model type and explanation constraints.
However, such approaches are limited to the techniques within the original framework, overlooking subsequent improvements and adaptations. This is significant, as standard, out-of-the-box techniques may be insufficient in scenarios requiring tailored solutions, such as applying LIME to audio data for audio sample explanations \cite{audiolime}. To address this, we have created the LIME overview page (\Cref{lime_webpage}), providing a quick overview of available techniques along with corresponding code implementations to facilitate efficient selection and application.

\textbf{Evaluation of LIME Techniques.}
The challenge of establishing widely adopted evaluation techniques for LIME-related approaches, as discussed in \Cref{lime_problems}, remains a significant concern. 
While a broad spectrum of suitable and well-established metrics already exists, their application is inconsistent across the community \cite{10.1145/3583558}. 
To address this inconsistency, developing a tailored evaluation framework specifically for LIME would be beneficial in assessing whether the identified issues are effectively mitigated. 
Such a framework could draw inspiration from existing XAI evaluation frameworks, which facilitate the comparison of fundamental metrics across various XAI techniques \cite{10.1007/978-3-031-63787-2_23, agarwal2024openxaitransparentevaluationmodel,klein2024navigatingmazeexplainableai}.

\textbf{Foundation Model Integration.}
In recent years, foundation models, such as large language models (LLMs), have gained widespread prominence. Trained on diverse datasets, these models encapsulate world knowledge by capturing the underlying structural patterns within the data. For instance, in reinforcement learning (RL) domains, foundation models are increasingly used during the agent exploration phase, leveraging the knowledge embedded in LLMs to enhance the agent’s performance \cite{xu2024languageagentsreinforcementlearning}.
This knowledge can also be utilized in XAI by assessing whether it aligns with the explained model’s decisions or by identifying biases in the world model through systematic cross-checking, enabling the deliberate integration of foundation models into explainability.
Another promising approach would be integrating a large language model (LLM) into LIME's feature generation and sample generation stages for text explanations. LLMs can enhance the locality of explanations by generating features that are more contextually aligned with the instance being explained. Furthermore, leveraging foundation models for feature generation could soften LIME’s strict locality, transitioning it toward a hybrid local-global explanation. This is because LLMs can naturally sample instances from the surrounding neighborhood, broadening the scope of the explanation. For sample generation, foundation models could guide the process by selectively sampling instances where LIME is uncertain, effectively using the LLM's knowledge as an exploration-driven function to improve the overall explanation quality. 

However, it is crucial to distinguish between scenarios where, for instance, an LLM is used to simulate a model's decision \cite{kroeger2024incontextexplainersharnessingllms} and those where a foundation model is directly incorporated into an explainability approach. In the latter case, the explanation is not generated by the foundation model itself; instead, its properties are incorporated into the explanation process. Research has already demonstrated this with LIME \cite{knab2024dseglime}, and similar approaches have been successfully applied to other explainability methods, such as SHAP \cite{sun2023explainconceptsegmentmeets}.

\textbf{Focus on Explainee.}
For whom are these techniques being developed? The most common motivation is to improve explanations for users seeking to understand the decision-making basis of an AI system within a given input. 
This assumption often suggests that the work applies to straightforward, user-centric scenarios. 
However, due to the frequent lack of accessible or complete public implementations, end users have limited or no opportunity to effectively benefit from the insights gained. Additionally, there can be a mismatch between how XAI techniques generate explanations for an instance and what the end user expects or finds useful \cite{freiesleben2023dearxaicommunityneed}.
The future directions for LIME-based research highlighted here emphasize the importance of involving end users more directly. This could include assisting them in selecting the appropriate LIME techniques, ensuring proper evaluation, or leveraging the contextual knowledge of foundation models. Therefore, this point is not a call for a specific new research direction but a recommendation for how future research should be structured. 

\subsection{Conclusion}

In this survey, we provide a comprehensive overview of XAI techniques built upon the widely used local and model-agnostic framework, LIME. We begin by describing the core principle of LIME: approximating model behavior in a localized region and illustrate how this approach adapts to various data modalities.
Building upon this foundation, we organize LIME’s underlying processes into a taxonomy, distinguishing the framework’s key subprocesses. We map the specific challenges each technique addresses, derived directly from the existing literature, to highlight the unique contributions of each method. Additionally, we examine LIME adaptations across different target modalities and application domains, emphasizing how the framework can be refined to enhance accuracy or adapt to specific domains.
Through this systematic analysis, we identify overarching trends within the research community and highlight persistent challenges that warrant further investigation. 

\vspace{1em} 
\noindent
\textbf{Acknowledgments.} This research was supported in part by the German Federal Ministry for Economic Affairs and
Climate Action of Germany (BMWK), and in part by the German Federal Ministry of Education
and Research (BMBF).
 
%
%
%
%
\bibliographystyle{splncs04}
\bibliography{bibtex}

\end{document}